\documentclass[9pt,conference]{IEEEtran}
\IEEEoverridecommandlockouts
\usepackage{cite}
\usepackage{amsmath,amssymb,amsfonts}
\usepackage{graphicx}
\usepackage{textcomp}
\usepackage{xcolor}

\usepackage{amsmath}
\usepackage{clrscode}
\usepackage{pifont}

\usepackage{multirow}
\usepackage{makecell}

\usepackage{microtype}
\usepackage{graphicx}
\usepackage{subfigure}
\usepackage{booktabs} 
\usepackage{amssymb,amsmath}
\usepackage{pifont}
\usepackage{color}
\usepackage[english]{babel}
\usepackage{array}
\usepackage{bm}
\usepackage{multirow}
\usepackage{enumitem}
\usepackage{pifont}
\usepackage{tikz}
\usepackage{amssymb}
\usepackage{enumitem}
\usepackage[english]{babel}
\usepackage{array}
\usepackage{bm}
\usepackage{multirow}
\usepackage{enumitem}
\usepackage{pifont}
\usepackage{graphicx}

\usepackage{caption}

\usepackage{lipsum}
\usepackage{ntheorem}
\newtheorem*{thm}{Theorem}

\usepackage{flushend}

\def\placeholder{NASAIC}

\def\BibTeX{{\rm B\kern-.05em{\sc i\kern-.025em b}\kern-.08em
    T\kern-.1667em\lower.7ex\hbox{E}\kern-.125emX}}
\begin{document}


\title{Co-Exploration of Neural Architectures and Heterogeneous ASIC Accelerator Designs Targeting Multiple Tasks}

\author{
\IEEEauthorblockN{
Lei Yang$^{1}$ \quad
Zheyu Yan$^{1}$ \quad
Meng Li$^{2}$ \quad
Hyoukjun Kwon$^{3}$ \quad
Liangzhen Lai$^{2}$ \quad
Tushar Krishna$^{3}$ \quad
Vikas Chandra$^{2}$ \quad \\
Weiwen Jiang$^{1,*\thanks{* W. Jiang is the corresponding author (wjiang2@nd.edu)}}$ \quad
Yiyu Shi$^{1}$ \quad
}
\IEEEauthorblockA{
\normalsize
$^{1}$ University of Notre Dame
$^{2}$ Facebook
$^{3}$ Georgia Institute of Technology
\\
wjiang2@nd.edu
}
}


\maketitle

\begin{abstract}
Neural Architecture Search (NAS) has demonstrated its power 
on various AI accelerating platforms such as Field Programmable 
Gate Arrays (FPGAs) and Graphic Processing Units (GPUs). However, 
it remains an open problem how to integrate 
NAS with Application-Specific Integrated Circuits (ASICs), despite 
them being the most powerful AI accelerating platforms. The major 
bottleneck comes from the large design freedom associated with 
ASIC designs. Moreover, with the consideration that multiple DNNs will run in parallel for different workloads with diverse layer operations and sizes, integrating heterogeneous ASIC sub-accelerators 
for distinct DNNs in one design can significantly boost performance, 
and at the same time further complicate the design space. 
To address these challenges, 
in this paper we build ASIC template set based on existing successful 
designs, described by their unique dataflows, so that the design space is significantly reduced. 
Based on the templates, we further propose a framework, namely \placeholder, which can simultaneously identify multiple DNN architectures and the associated heterogeneous ASIC 
accelerator design, such that the design specifications (specs) can be satisfied, while the accuracy can be maximized.
Experimental results show that compared with successive NAS and ASIC design optimizations which lead to design spec violations, \placeholder~can guarantee the results to meet the design specs with 17.77\%, 2.49$\times$, and 2.32$\times$ reductions on latency, energy, and area and with 0.76\% accuracy loss. To the best of the
authors' knowledge, this is the first work 
on neural architecture and ASIC accelerator design co-exploration. 
\end{abstract}


\setlength{\textfloatsep}{3pt}
\setlength{\floatsep}{1pt}
\setlength{\dbltextfloatsep}{3pt}
\setlength{\belowdisplayskip}{3pt}
\setlength{\abovedisplayskip}{3pt}

\section{Introduction}
\label{s:Introduction}


Recently, Neural Architecture Search (NAS) \cite{zoph2016neural,liu2018darts,pham2018efficient} successfully opens up the design freedom to automatically identify the neural architectures with the maximum accuracy; in addition, hardware-aware NAS \cite{wu2018fbnet,cai2018proxylessnas,jiang2019accuracy,hao2019fpga,jiang2019hardware,jiang2019device,jiang2019achieving,yang2020coexplore,lu2019neural,zhang2019neural,bian2020nass} further enables the hardware design space to jointly identify the best architecture and hardware designs in maximizing network accuracy and hardware efficiency. 
Most of the existing hardware-aware NAS approaches focus on GPUs or 
Field Programmable Gate Arrays (FPGAs).

\begin{figure}[t]
  \centering
  \includegraphics[width=3.3889 in]{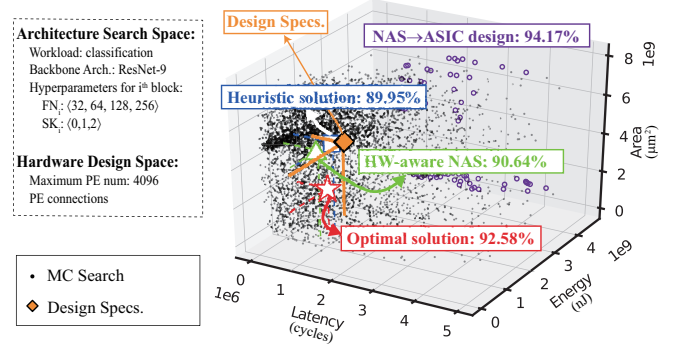}
  \caption{Neural architecture search space and hardware design space exploration: solutions from 
  successive NAS and ASIC design; solution from NAS in aware of an ASIC design; the closest-to-spec solution; and the optimal solution from 10,000 Monte Carlo (MC) runs. (Best viewed in color)}
  \label{fig:motivation}
\end{figure}

On the other hand, among all AI accelerating platforms, application-specific integrated circuits (ASICs), composed 
of processing elements (PEs) connected in different topologies, can
provide incomparable energy efficiency, latency, and 
form factor~\cite{chen2016eyeriss,parashar2017scnn,xu2018scaling}. 
Most 
existing ASIC accelerators, however, target 
common neural architectures  \cite{du2015shidiannao,2017NVDLA,chen2016eyeriss} and do not reap 
the power of NAS. Though seemingly straightforward, integrating NAS with ASIC designs is not a 
simple matter, as can be seen from the image classification 
example in  Fig.~\ref{fig:motivation}. The neural architecture 
search space is formed by 
ResNet9 \cite{2019ResNet9} 
with adjustable hyperparameters. The hardware design space is formed by ASICs with adjustable number of PEs and their connections. 
The results are depicted in a three-dimensional space, where the three axes represent different hardware metrics and each point represents a solution of paired neural architecture and ASIC design. 
From the figure 
we can see that when NAS and ASIC design are performed successively, 
all the solutions (denoted by circles) violate user-defined 
hardware design specifications (design specs, denoted by diamond). When NAS is done in aware
of a particular ASIC design, the resulting solution (denoted by triangle)
has lower accuracy 
compared with the optimal one (denoted by star) from 10,000 Monte Carlo
runs, which uses a different ASIC design. 
A simple heuristic to pick a solution with latency, 
energy and area closest to the design specs (denoted by square) would also be sub-optimal. 
It is therefore imperative to jointly explore the neural architecture search space and hardware design space to identify the optimal solution.

However, such a task is quite challenging, 
primarily due to the 
large design space of ASICs where a same set of PEs 
can constitute numerous topologies (and thus dataflows). Enumeration 
is simply out of the question. 
In addition, when ASIC accelerators are deployed on the edge, 
they usually need to handle 
multiple tasks involving multiple DNNs.
For instance, tasks like object detection, image segmentation, and classification can be triggered simultaneously on augmented reality (AR) glasses \cite{2019glasses}, each of which relies on one kind of DNN.
Since the DNNs for different tasks can have distinct architectures, one dataflow cannot fit all of them; meanwhile, multiple tasks need to be 
executed concurrently, which requires task-level parallelism.
As such, it is best to integrate multiple heterogeneous sub-accelerators (corresponding to different dataflows) 
into one accelerator to improve performance and energy efficiency, which has been verified in \cite{kwon2019herald}. Yet this further complicates the design space.

To address these challenges, in this paper, 
we establish a link between NAS and ASIC accelerator design. 
Instead of a full-blown exploration of the design space, we 
observe that there already exist a few 
great ASIC accelerator designs such as 
Shidiannao \cite{du2015shidiannao}, NVDLA \cite{2017NVDLA}, and Eyeriss \cite{chen2016eyeriss}. Each of these designs 
has its unique dataflow, and the accelerator is determined 
once the hardware resource associated with the dataflow is given. 
As such, we can create 
a set of ASIC templates, where each template corresponds to one specific dataflow, so that the design space can be significantly narrowed down to the selection of templates to form a heterogeneous accelerator, and the allocation of hardware resources (e.g., the number of PEs and NoC bandwidth) to the selected templates.

Based on the template concept, we then further propose a neural architecture and ASIC design co-exploration framework, namely \placeholder, for multiple tasks targeting edge devices. The objective of \placeholder~is to identify the best neural architectures for each task and the ASIC design, such that all design specs can be met while the accuracy of the neural architectures can be maximized. 
Specifically, we devise a novel controller that can simultaneously predict hyperparameters of multiple DNNs together with the parameters of hardware resource allocation for different template selections.
Based on the state-of-the-art cost model~\cite{kwon2018maestro}, we separately explore the mapping and scheduling of neural architectures onto ASIC templates.
Finally, a reward is generated to update the controller.
To accelerate the search process, we apply the early pruning technique to remove neural architectures that cannot satisfy design specs without training. 
Experimental results on the workload with the mixed classification and segmentation tasks show that, compared with solutions generated by the successive NAS and ASIC design optimization which cannot satisfy the design specs, those from \placeholder~can guarantee to meet the design specs with 17.77\%, 2.49$\times$, and 2.32$\times$ reductions in latency, energy, and area and with only 0.76\% average accuracy loss on these two tasks. 
Furthermore, compared with hardware-aware NAS for a fixed ASIC
design, \placeholder~can achieve 3.65\% higher accuracy. To the best of the authors' knowledge, this is the first 
work on neural architecture and 
ASIC design co-exploration.

\section{Background and Challenges}
\label{s:Background}


We are now witnessing the rapid growth of NAS.
Since the very first work for NAS with reinforcement learning \cite{zoph2016neural}, there has been tremendous work to study efficient neural architecture search \cite{liu2018darts,pham2018efficient}.
Integrating hardware awareness in the search loop opens a new research direction, which attracts research efforts on hardware-aware NAS \cite{wu2018fbnet,cai2018proxylessnas}.
Taking one step further, most recently, co-exploration of neural architecture and hardware design is proposed \cite{jiang2019accuracy,hao2019fpga}.
Unlike the original NAS with mono-objective on maximizing accuracy, those hardware-aware NAS frameworks take inference latency 
into consideration, and push 
forward the deployment of DNNs on edge devices.
NAS has been applied to 
GPUs and FPGAs but not yet to ASICs, though ASICs are the most efficient ones among all AI accelerating 
platforms \cite{zhang2018thundervolt,zhang2019compact}. 

Two so-far-unseen but urgent-to-solve challenges exist.






\textit{\textbf{Challenge 1:} How to enable the co-exploration of neural architectures and ASIC accelerator designs?}

The large design space of ASIC accelerators hinders the application of NAS to ASIC accelerators.
Unlike GPUs with fixed hardware or FPGAs with well-structured  hardware, ASIC designs grant the maximum flexibility to designers to determine the hardware organization.
This enables to pursue the maximum efficiency; however, it significantly enlarges the design space.
Fortunately, there exist extensive research works in designing ASIC AI accelerators \cite{du2015shidiannao,2017NVDLA,chen2016eyeriss}, making it possible to shrink the design space on top of existing designs.

 
Among all ASIC accelerator designs, one of the key observations is that each design has a specific dataflow, such as Shidiannao \cite{du2015shidiannao}, NVDLA \cite{2017NVDLA} , and Eyeriss \cite{chen2016eyeriss} styles.
For instance, NVDLA \cite{2017NVDLA} involves an adder-tree to calculate the partial sum of output feature maps. 
Inspired by this, we propose to build a set of accelerator templates, each of which has a dataflow style, resulting in a fixed hardware structure. 
On top of it, we only need to allocate resource for templates, without changing hardware structures.
Consequently, the design space can be significantly shrunk, and in turn, it enables the co-exploration of neural architectures and ASIC designs by incorporating hardware allocation parameters.


\textit{\textbf{Challenge 2:} Multiple neural architectures need to be identified under the unified design specs.}

Another challenge is that the realistic applications on edge devices require the collaboration of multiple tasks, which involves multiple DNNs.
In addition, all these DNNs will be executed on the accelerator with unified design specs, including latency, energy, and area.
In consequence, sequentially optimizing each DNN using hardware-aware NAS will not work; instead, the multiple neural architectures need to be 
simultaneously optimized under the unified design specs.


Integrating multiple DNNs in one accelerator brings one further challenge.
DNNs for different tasks have distinct architectures,
yet one dataflow is not suitable for all architectures. 
For instance, NVDLA style \cite{2017NVDLA} ($DF_2$ in Fig.~\ref{fig:problem}) loads one pixel from each activation channel for one computation.
In order to fully use the computation resource, it favors convolution layers with large activation channel but low activation resolution; while Shidiannao style \cite{du2015shidiannao} ($DF_1$ in Fig.~\ref{fig:problem}) is on the opposite.
As a result, NVDLA style works better for ResNets, 
while Shidiannao works better for U-Nets. 
As demonstrated in \cite{kwon2019herald}, we can 
integrate multiple heterogeneous sub-accelerators using 
a network-on-chip topology through Network Interface Controller (NIC) in one AISC accelerator, which further complicates the design space. 

In this work, we will address the above challenges.

\section{Problem Definition}
\label{s:Problem Definition}

In this section we will first define multi-task workloads and heterogeneous accelerators, and then formulate the problem of neural architecture and ASIC design co-exploration.


\begin{figure}[t]
  \centering
  \includegraphics[width=3.3967 in]{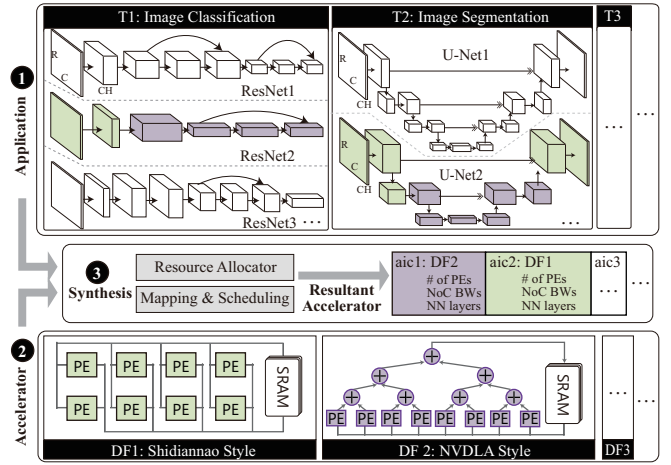}
  \caption{Overview: co-exploration with three layers of optimizations.}
  \label{fig:problem}
\end{figure}

Fig.~\ref{fig:problem} demonstrates an overview of the co-exploration, which involves three exploration layers: {\large\ding{202}} ``Application'', {\large\ding{203}} ``Accelerator'', and {\large\ding{204}} ``Synthesis''.
The application layer determines the neural architectures to be applied, while the accelerator layer creates the an ASIC template set based on the dataflow style of the existing accelerator designs.
Acting as the bridge, the synthesis layer allocates a template together with the resources to each sub-accelerator, then maps and schedules the network layers to sub-accelerators.
In the following text, we will define each exploration layer in detail.

\begin{figure}[t]
  \centering
  \includegraphics[width=3.4399 in]{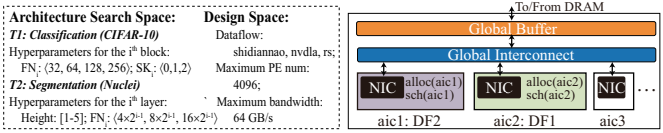}
  \caption{Left: search spaces for both NAS and ASIC accelerator designs. Right: the resultant heterogeneous ASIC accelerator. }
  \label{fig:NOC}
\end{figure}

{\large\ding{202}} \textbf{Application.} The application workload considered in this work has multiple AI tasks which involve a DNN model for each task.
A workload with $m$ tasks is defined as $W=\langle T_1,T_2,\cdots,T_m\rangle$. 
Fig.~\ref{fig:problem} shows an example with two tasks (i.e., $T_1$ for classification and $T_2$ for segmentation).
Task $T_i\in W$ corresponds to a DNN architecture $D_i$, which forms a set $D$ with $m$ DNN architectures.
We define a DNN architecture as $D_i=\langle B_i, L_i, H_i,acc_i\rangle$.
$D_i$ is composed of a backbone architecture $B_i$, a set of layers $L_i$, a set of hyperparameters $H_i$, and an accuracy $acc_i$. 
For example, in Fig.~\ref{fig:problem}, backbone architecture $B_1$ for classification task $T_1$ is ResNet9 \cite{2019ResNet9}, and its hyperparameters include the number of filter  ($FN$) and the number of skip layers ($SK$) for each residual block, as shown in Fig.~\ref{fig:NOC} (left); while for $T_2$, backbone architecture $B_2$ is U-Net \cite{ronneberger2015u} whose hyperparameters include the height ($Height$) and filter numbers ($FN$) for each layer.

Based on the above definition, we define the neural architecture search function $H_i=nas(D_i)$, which determines hyperparameters $H_i$ in DNN $D_i$ to identify one neural architecture. Note that NAS \cite{zoph2016neural} is to determine $nas(D_i)$ with the mono-objective of maximizing accuracy $acc_i$.
As shown in Fig. \ref{fig:problem}, each set of hyperparameters corresponds to one neural architecture, and we determine $nas(D_i)$ to identify a specific neural architecture for task $T_i$ (colored ones).

{\large\ding{203}} \textbf{ASIC Accelerator.}
A heterogeneous ASIC accelerator formed by multiple sub-accelerators connected in a NoC topology through NIC is shown in Fig.~\ref{fig:NOC} (right).
Define $AIC=\langle aic_1, aic_2,\cdots aic_k\rangle$ to be a set of $k$ sub-accelerators.
A sub-accelerator $aic_i=\langle df_i, pe_i,bw_i \rangle$ has three properties: dataflow style $df_i$, the number of PEs $pe_i$, and the NoC bandwidth $bw_i$.
With a set of predefined dataflow templates to choose from, as shown in Fig. \ref{fig:problem}, the ASIC design space is significantly narrowed down from choosing specific unrolling, mapping and data reuse patterns to allocating resources (one template with associated PEs and bandwidth) to each sub-accelerator.
Kindly note that according to the template and the mapped network layers, the memory size can be determined to support the full use of hardware, as in \cite{kwon2018maestro}.
Therefore, memory size will not be explored in the search space.


{\large\ding{204}} \textbf{Synthesis.} Based on the definition of applications and accelerators, next, we present the synthesis optimization.

\textit{Resource allocation.} On the hardware side, we design each sub-accelerator in set $AIC=\langle aic_1, aic_2,\cdots aic_k\rangle$, given a set of dataflow templates $DF=\langle DF_1, DF_2, \cdots DF_q\rangle$, the maximum number of PEs (e.g., $NP= 4096$) and the maximum bandwidth (e.g., $BW 64GB/s$.
Note that since $DF$ contains different dataflows, the resultant accelerator will be heterogeneous if more than one type of dataflows are mapped to $AIC$.
By reducing the size of $DF$ to one, the proposed techniques can be used for homogeneous designs.


We define an allocation function $alloc(aic_i)$ to determine the dataflow template from $DF$, and the PEs and bandwidth used for $aic_i$, such that $\sum\nolimits_{i=1\cdots |AIC|}\{pe_i\}\le NP$ and $\sum\nolimits_{i=1\cdots |AIC|}\{bw_i\}\le BW$.
As an example, Fig.~\ref{fig:problem} illustrates two kinds of dataflow templates: shidiannao \cite{du2015shidiannao} and NVDLA \cite{2017NVDLA}.
The resultant accelerator (in Fig.~\ref{fig:problem} {\large\ding{204}}) is composed of two heterogeneous sub-accelerators with different dataflow templates, PE numbers and bandwidth.

\textit{Mapper and scheduler.} On the software side, we map network layers to sub-accelerators and determine their execution orders on each sub-accelerator.
A map function $map(l_{i,j})=aic_k$ is defined, which indicates the $j^{th}$ network layer $l_{i,j}$ in the $i^{th}$ DNN $D_i$ to be mapped to the $k^{th}$ sub-accelerator $aic_k$.
Based on the mapping, we determine the execution order of network layers on sub-accelerator $aic_k$ following a schedule function $sch(aic_k)$.

The synthesis results can be evaluated via four metrics, including accuracy, latency, energy, and area. 
In this work, we aim to maximize the accuracy of DNNs under the given design specs on latency ($LS$), energy ($ES$) and area ($AS$).

\noindent\textbf{Problem Definition.}
Based on all the above definitions, we formally define the optimization problem as follows: given a multi-task workload $W$, the backbone neural architecture for each DNN in set $D$, a set of sub-accelerators $AIC$, a set of dataflow templates $DF$, the maximum number of PEs and bandwidth, and design specs ($LS$, $ES$, $AS$), we determine:
\begin{itemize}
    \item $nas(D_i)$: architecture hyperparameters of each DNN $D_i\in D$;
    \item $alloc(aic_k)$: the dataflow and resource allocation for each sub-accelerator $aic_k\in AIC$;
    \item $map(l_{i,j})$ and $sch(aic_k)$: the mapping of network layers to sub-accelerators and their schedule orders; 
\end{itemize}
such that the maximum accuracy of DNNs can be achieved while all design specs and resource constraints are met; i.e., $max=weighted(D)$, $s.t.$, $rl\le LS$, $re\le ES$, $ra\le AS$ $\sum\nolimits_{i=1\cdots |AIC|}\{pe_i\}\le NP$, $\sum\nolimits_{i=1\cdots |AIC|}\{bw_i\}\le BW$, where $rl,re,ra$ represent latency, energy, and area of the resultant accelerator, and a $weighted$ function defined in next section is to get the accuracy of all networks, which can be functions like $avg$ (maximize the average accuracy) or $min$ (maximize the minimum accuracy).

\section{Proposed co-exploration framework: \placeholder}
\label{s:Framework}

\begin{figure}[t]
  \centering
  \includegraphics[width= 3.3 in]{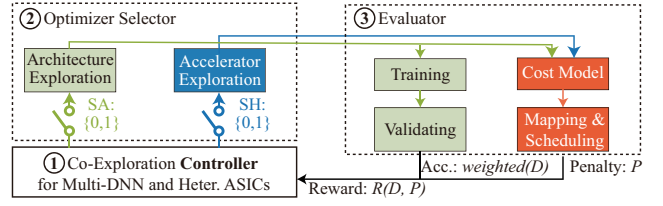}
  \caption{\placeholder: parameters for neural architecture and accelerator are first determined by controller; then the identified neural architecture and accelerator will be evaluated; finally, a reward will be generated by the evaluation results to feedback and update the controller.}
  \label{fig:framework}
\end{figure}

This section will present the details of \placeholder~that addresses 
the problem formulated in Section~\ref{s:Problem Definition}.
Fig.~\ref{fig:framework} demonstrates the overview of \placeholder.
It contains three components, including {\large\ding{172}} controller, {\large\ding{173}} optimizer selector, and {\large\ding{174}} evaluator.
In general, the controller samples neural architectures and hardware resource allocation in each episode (aka. iteration). Then the predicted sample goes through the optimizer selector and evaluator to generate the accuracy and hardware cost. Finally, a reward is generated to update the controller.
All the components work together to generate solutions with high weighted accuracy and meet all design specs. 
To illustrate \placeholder~framework, we apply reinforcement learning approach in this paper. 
Based on the formulated reward function, other optimization approaches, such as evolution algorithms, can also be applied. Note that 
since the hardware constraints are non-differentiable, differentiable neural architecture search (DARTS) cannot be applied.
In the following text, we will introduce each component in detail.

{\large\ding{172}} \textbf{Multi-Task Co-Exploration Controller.} The controller is the key component in \placeholder.
Driven by the requirement of multi-task in one application workload, we propose a novel reinforcement-learning based Recurrent Neural Network (RNN) controller to simultaneously predict multiple neural architectures. 
In addition, we integrate accelerator design parameters into the controller to realize a genuine co-exploration of neural architectures and hardware designs.

\begin{figure}[t]
  \centering
  \includegraphics[width= 3.471 in]{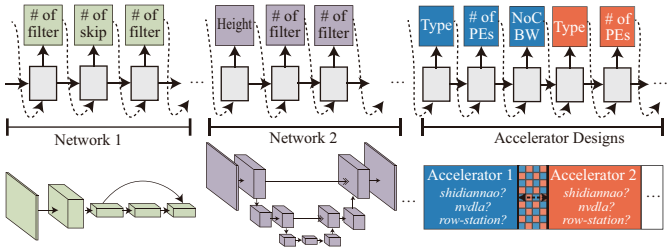}
  \caption{Co-exploration controller for multiple tasks: determine neural architecture hyperparameters, and hardware design parameters.}
  \label{fig:controller}
\end{figure}

Fig.~\ref{fig:controller} demonstrates the proposed controller.
It is composed of $N$ segments, where $N$ is the sum of task number in workload $W=\{T_1,T_2,\cdots,T_m\}$ and sub-accelerator number in set $AIC=\{aic_1,aic_2,\cdots,aic_k\}$; i.e., $N=m+k$. 
The first $m$ segments correspond to $m$ DNNs, while the remaining segments correspond to $k$ sub-accelerators. 
For the segment associated with a DNN, say $D_i\rangle$, its outputs determine $D_i$'s hyperparameters, i.e., the $nas(D_i)$ function.
For instance, in Fig.~\ref{fig:controller}, the first segment predicts the filter numbers (FN) and skip layers (SK).
Similarly, the segment for sub-accelerator $aic_k$ determines its hardware design parameters, i.e., the $alloc(aic_k)$ function, as shown in the right part of Fig.~\ref{fig:controller}.

We employ reinforcement learning method to update the controller and predict new samples.
Specifically, in each episode, the controller first predicts a sample, and gets its reward $R$ based on the evaluation results form components {\large\ding{174}} and {\large\ding{175}}.
Then, we employ the Monte Carlo policy gradient algorithm \cite{williams1992simple} to update the controller:
\begin{equation}
    \nabla J(\theta) = \frac{1}{m}\sum_{k=1}^{m}\sum_{t=1}^{T}\gamma^{T-t}\nabla_{\theta}\log \pi_{\theta} (a_{t}|a_{(t-1):1})(R_{k}-b)
\end{equation}
where $m$ is the batch size and $T$ is the number of steps in each episode. Rewards are discounted at every step by
an exponential factor $\gamma$ and the baseline $b$ is the average exponential moving of rewards.

{\large\ding{173}} \textbf{Optimizer Selector.} We integrate an optimizer selector in \placeholder~to accelerate the search process. 
This is based on the observation that the speed of hardware evaluation is much faster than the training process. 
Specifically, as shown in Fig.~\ref{fig:framework}, we add two switches ($SA$ for neural architecture exploration and $SH$ for hardware design exploration).
In terms of the status of switches, the framework can perform different functions listed as follows:
\begin{itemize}[noitemsep,topsep=0pt,parsep=0pt,partopsep=0pt]
  \item $SA=1,SH=0$, it performs conventional NAS, like \cite{zoph2016neural}. 
  \item $SA=0,SH=1$, it uses the previous neural architecture and explores hardware designs only. In this case, we aim to obtain valid accelerator design for the neural architecture, and therefore, we do not consider the accuracy in reward.
  \item $SA=1,SH=1$, it predicts new neural architectures and hardware designs.
\end{itemize}
\placeholder~repeatedly conducts the following two steps $\beta$ times: (1) both $SA$ and $SH$ are closed for 1 step, aiming to obtain new neural architecture and hardware design; (2) the switch $SA$ is opened for $\phi$ steps, in order to explore the best hardware for a previous identified neural architecture.
Kindly note that the first step is carried out in a non-blocking scheme, such that one training and $\beta$ times hardware exploration can be conducted in parallel.
Once all hardware explorations are completed and no feasible hardware design is found, it will terminate the training process to accelerate the search process. 

{\large\ding{174}} \textbf{Evaluator.} The evaluator contains two paths: (1) via the training and validating to obtain networks' accuracy; (2) via cost modeling, mapping and scheduling to generate penalty in terms of design specs.

\textit{Training and validating}
In this path, hyperparameters $H_i$ for DNN architecture $D_i$ are obtained from controller.
For each DNN $D_i\in D$, we train it from scratch and get its accuracy $acc_i$ on a held-out validation dataset.
Based on the accuracy, we obtain the weighted accuracy $weighted(D)$ for calculating the reward $R$ as follows:
\begin{equation}\label{equ:acc}
    weighted(D) = \sum\nolimits_{i=1,2,\cdots,|W|}\{\alpha_i\times acc_i\}
\end{equation}
where $|W|$ is the total number of tasks in the given workload, and $\alpha_i$ is a weight ranging from 0 to 1, such that $\sum\nolimits_{i=1,2,\cdots,|W|}\{\alpha_i\}=1$.



\textit{Mapping and scheduling} 
On this path, a set of identified DNN architectures $D$ and a set of determined sub-accelerator $AIC$ are given by controller.
We need to get the hardware metrics including latency $rl$, energy $re$, and area $ra$.
\placeholder~incorporates the state-of-the-art cost model, MAESTRO \cite{kwon2018maestro}, and a mapping and scheduling algorithm to obtain the above metrics.
For area $ra$, we can directly obtain it from MAESTRO with the given sub-accelerator $AIC$.
The latency $rl$ and energy $re$ are determined by the mapping and scheduling.
To develop an algorithm for mapping and scheduling, we need to obtain the latency and energy of each layer on different sub-accelerators.
Let $L=\bigcup\nolimits_{D_k\in D}\{L_k\}$ be the layer set.
For a pair of network layer $\forall l_i\in L$ and sub-accelerator $aic_j\in AIC$, we can input them to MAESTRO to get the latency $l_{i,j}$ and energy $e_{i,j}$.


The problem can be proved to be equivalent to the traditional heterogeneous assignment problem \cite{ito1998ilp,shao2005efficient}: given the latency $l_{i,j}$ and energy cost $e_{i,j}$ for each layer $i$ on sub-accelerator $j$, the dependency among layers, and a timing constraint $LS$, we are going to determine the mapping and scheduling of each layer on one sub-accelerator, such that the energy cost $re$ is minimized while the latency $rs\le LS$. We denote $HAP$ to be an optimal solver, i.e., $re=HAP(D,AIC,LS)$. Then, we have the following theorem.
\begin{thm}
Given a layer set $D$, a sub-accelerator set $AIC$, and design specs on latency $LS$ and energy $ES$, the design specs can be met if and only if $re=HAP(D,AIC,LS)\le ES$. 
\end{thm}

\begin{figure*}[t]
  \centering
  \includegraphics[width= 7 in]{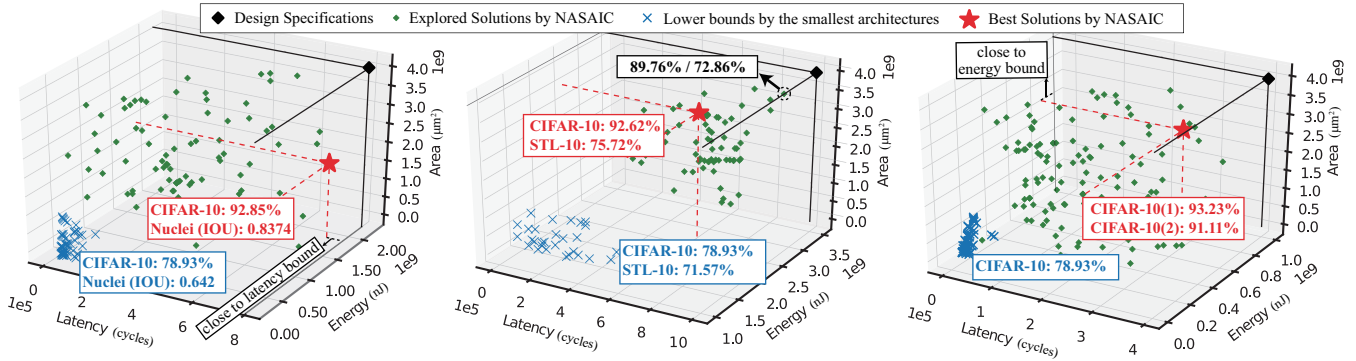}
  \vspace{-10pt}
  \caption{Exploration results obtained by \placeholder~for three different workloads under design specs: (left) $W1$ with CIFAR-10 and STL-10 datasets; (middle) $W2$ with CIFAR-10 and Nuclei;  (right) $W3$ with CIFAR-10 dataset. (Best viewed in color)}
  \label{fig:expDse}
\end{figure*}

The above theorem can be proved using contradiction. Due to the space limitation, the detailed proof is omitted.
Based on this theorem, the latency $rl$ and energy $re$ are obtained by the solver $HAP$, which can be instantiated by Integer-Linear Programming (ILP) for the optimal solution; however, since ILP is time-consuming, this paper applies a heuristic approach in \cite{shao2005efficient} to accelerate the search process. 
On top of the obtained hardware metrics and the given design specs, we formulate a penalty function.
Penalty is determined in terms of the degree that the solution exceeds the design specs, and no penalty if all design specs are met, which is formulated as follows:
\begin{equation}
    P=\frac{\max(rl-LS,0)}{(bl-LS)}+\frac{\max(re-ES,0)}{(be-ES)}+\frac{\max(ra-AS,0)}{(ba-AS)}
\end{equation}
where $bl$, $be$, $ba$ are the upper bounds for the metrics, which can be obtained by exploring the hardware design space using the neural architecture identified by NAS, as the circles in Fig.~\ref{fig:motivation}.



Finally, based on all the above evaluation results, we calculate the reward with a scaling variable $\rho$, listed as follows:
\begin{equation}
    R(D,P)=weighted(D)-\rho\times P
\end{equation}

\section{Experimental Evaluation}
\label{s:Experiment}

We evaluate the efficacy of the proposed framework, \placeholder, using different application workloads and hardware configurations.
Results reported in this section demonstrate that \placeholder~can efficiently identify accurate neural architectures together with AISC accelerator designs that are guaranteed to meet the given design specs, while achieving high accuracy for multiple AI tasks.


\noindent\textit{A. Evaluation Environment}

\noindent\textbf{Application workloads:}
We use typical workloads on AR glasses in applications such as driver assistance or augmented medicine to demonstrate the efficacy of NASAIC. 
In these workloads, the core tasks involve classification and segmentation, where 
representative datasets such as CIFAR-10, STL-10, and 
Nuclei are commonly employed, along with light-weight 
neural architectures. We synthesize the following 
three workloads.

\begin{itemize}
    \item W1: Tasks on one classification dataset (CIFAR-10) and one segmentation dataset (Nuclei).
    \item W2: Tasks on two classification datasets (CIFAR-10, STL-10).
    \item W3: Tasks on the same classification dataset (CIFAR-10).
\end{itemize}


The backbone architectures and their search space for the 
above tasks are defined as follows.
For the classification tasks, we select ResNet9 \cite{2019ResNet9}, which contains multiple residual blocks, as the architecture backbone.
During NAS, the number of convolution layer and the number of filter channels for each residual block are searched and then determined.
For CIFAR-10, we employ 3 residual blocks, and parameter options for each block are depicted in Fig.~\ref{fig:motivation}(a); while for STL-10, considering that its input images have higher resolution (i.e., $96\times 96$ pixels), we deepen the network to 5 residual blocks, and increase the maximum number of convolution layers in each residual block to 3 and the maximum number of filter channel to 512 for each block.
For the segmentation tasks, we use U-Net \cite{ronneberger2015u} as the architecture backbone. 
The search space for this backbone architecture includes the number of height and filter channel number in each layer, as shown in Fig.~\ref{fig:motivation}.
Note that we follow the standard NAS approach \cite{zoph2016neural} to hold out a part of data from training images to be the validation set, and the training parameters (e.g., batch size, learning rate, and etc.) follow ResNet9 \cite{2019ResNet9} and U-Net \cite{ronneberger2015u}.

\noindent\textbf{Hardware configuration:}
Accelerator design includes the allocation of hardware resources to sub-accelerators, and the selection of dataflow for each sub-accelerator.
For resource allocation, we set the maximum number of PEs to be 4096 and the maximum NoC bandwidth to be 64GB/s, in accordance to \cite{kwon2019herald}.
Note that, our proposed \placeholder~can support arbitrary number of sub-accelerators; for simple demonstration, we make a case study by integrating two sub-accelerators.
Specifically, each sub-accelerator uses one of the following dataflows: Shidiannao (abbr. shi) \cite{du2015shidiannao}, NVDLA (abbr. dla) \cite{2017NVDLA}, and row-stationary \cite{chen2016eyeriss} style.
In the case where one sub-accelerator has no resource allocation, the design degenerates to a single large accelerator; while in the case where sub-accelerators have exactly the same allocation, the design degenerates to homogeneous accelerators.

 Hardware constraints on latency, energy and area will be set by designers (users), according to their own use cases.
To evaluate the effectiveness of \placeholder, we set distinct and strict design specs,  including \textbf{L}atency ($cycles$), \textbf{E}nergy ($nJ$), \textbf{A}rea ($\mu m^2$), for each application workload as follows: 
$\langle 8e5, 2e9, 4e9\rangle$ for $W1$; $\langle 1e6, 3.5e9, 4e9\rangle$ for $W2$; $\langle 4e5, 1e9, 4e9\rangle$ for $W3$.


\noindent\textbf{\placeholder~setting:}
For exploration parameters, we set $\beta=500$ and $\phi=10$, indicating that we explore the search space for 500 episodes and 10 accelerator designs in each episode.
For reward calculation parameters, we set $\alpha_1=\alpha_2=0.5$ to calculate the weighted accuracy, and $\rho=10$.
Controller RNN is trained by RMSProp optimization, with the initial learning rate of 0.99 and exponential decay of 0.5 for 50 steps.
All experiments are conducted on a server with a 48-thread Intel Xeon CPU and one NVIDIA Tesla P100 GPU.
NASAIC only takes around 3.5 GPU Hours to complete the exploration 
for each workload, which mainly benefits from the early pruning from optimizer selector component in \placeholder~(see Section \ref{s:Framework} {\large\ding{173}}).



    

\noindent\textit{B. Design Space Exploration}

Fig.~\ref{fig:expDse} demonstrates the exploration results of \placeholder~on three application workloads.
In this figure, the x-axis, y-axis, and z-axis represent latency, energy and area, respectively.
The black diamond indicates the design specs (upper bound); each green diamond is a solution (neural architecture-ASIC design pair) explored by \placeholder; each blue cross is a solution based on the smallest neural network in the search space combined with different ASIC designs (lower bound); and the red star refers to the best solution in terms of the average accuracy explored by \placeholder.
The numbers in the rectangles with blue, green, and red colors represent the accuracy of the smallest network, the inferior solutions, and our best solutions, respectively. 

We have several observations from Fig.~\ref{fig:expDse}. 
First, \placeholder~can guarantee that all the explored solutions meet the design specs.
Second, the identified solutions have high accuracy. The accuracy on CIFAR-10 of the four solutions are 92.85\%, 92.62\%, 93.23\%, and 91.11\%, while the accuracy lower bounds from the smallest network is 78.93\%.
Similarly, for STL-10, the accuracy is 75.72\% compared with the lower 
bound of 71.57\%. For Nuclei, the IOU (Intersection Over Union) is 
0.8374 compared with the lower bound of 0.6462.
Third, we observe that the best solutions of $W1$ and 
$W3$ identified by \placeholder~are quite close to the boundary defined by one of the three design specs, which indicates that in these cases the 
accuracy is bounded by resources.
For $W1$, the energy of the identified solution is 97.12\% of the spec; while for $W3$, the latency of the identified solution is 93.4\%.
This gives designers insights on if/where the hardware bottleneck is that prevents the accelerator from getting higher accuracy, and thus they can loose such constraint to increase the accuracy if necessary.
On the other hand, for $W2$ (middle of Fig. \ref{fig:expDse}), our best solution is farther away from the specs compared with solution $S$ pointed out by the arrow ($S$ is one of the explored solutions by \placeholder).
However, the accuracy of $S$ for CIFAR-10 and STL-10 are 2.86\% and 2.91\% lower than the best solution.
This reflects that the best solution may not always be the one closest to the specs, and therefore, heuristics that select the solution that 
is closest to the specs cannot work. 

\noindent\textit{C. Results on Multiple Tasks for Multiple Datasets}

    
    
    
\captionsetup{margin=0pt,justification=justified}

\begin{table}[t]
  \scriptsize
  \centering
  \tabcolsep 1.1pt
  \renewcommand\arraystretch{1.3}
  \caption{Comparison between successive NAS and ASIC design (NAS$\rightarrow$ASIC), ASIC design followed by hardware-aware NAS (ASIC$\rightarrow$HW-NAS), and \placeholder.}
  \label{tab:resMulti}
  \begin{tabular}{c|c|c|c|c|c|c|c}
    \hline
    Work. & Approach    & Hardware & Dataset  &  Accuracy & L /{\tiny $cycles$} & E /{\tiny $nJ$} & A /{\tiny $\mu m^2$}  \\
    \hline 
    \multirow{6}{*}{W1} & \multirow{2}{*}{NAS$\rightarrow$ASIC} & \multirow{2}{*}{\thead{$\langle dla,2112,48 \rangle$ \\ $\langle shi,1984,16 \rangle$}}
 & CIFAR-10 &  94.17\%  & 9.45e5&	3.56e9 &	4.71e9 \\
    
    &  &  & Nuclei   &  83.94\%  & $\times$ & $\times$ & $\times$ \\      
    \cline{2-8}
    & ASIC$\rightarrow$ & \multirow{2}{*}{\thead{$\langle dla,1088,24 \rangle$ \\ $\langle shi,2368,40 \rangle$}} & CIFAR-10 &  91.98\%  & 5.8e5
& 1.94e9 & 3.82e9\\
    
    &  HW-NAS &  & Nuclei   &  83.72\%  & {$\checkmark$}
&  {$\checkmark$} &  {$\checkmark$}\\        
    \cline{2-8}
    & \multirow{2}{*}{\placeholder}   & \multirow{2}{*}{\thead{$\langle dla,576,56 \rangle$ \\ $\langle shi,1792,8 \rangle$}}  & CIFAR-10 &  92.85\% &  {\textbf{7.77e5}} & {\textbf{1.43e9}} & {\textbf{2.03e9}}\\
    & & & Nuclei   &  83.74\% & {$\checkmark$} & {$\checkmark$} & {$\checkmark$} \\
    \hline
  
    \hline
    \multirow{6}{*}{W2} &\multirow{2}{*}{NAS$\rightarrow$ASIC} & \multirow{2}{*}{\thead{$\langle dla,2368,56 \rangle$ \\ $\langle shi, 1728, 8 \rangle$}}
 & CIFAR-10 &  94.17\%  & {9.31e5} & 3.55e9 & 4.83e9 \\
    
    &  &  & STL-10   &  76.50\%  & {$\checkmark$} & $\times$ & $\times$ \\      
    \cline{2-8}
    & ASIC$\rightarrow$ & \multirow{2}{*}{\thead{$\langle dla,2112,24 \rangle$ \\ $\langle shi, 1536, 40 \rangle$}}  & CIFAR-10 &  92.53\%  & 9.69e5 & 2.90e9 & {3.86e9} \\
    
    & HW-NAS &   & STL-10   &  72.07.\%  & {$\checkmark$} & {$\checkmark$} & {$\checkmark$} \\        
    \cline{2-8}
    & \multirow{2}{*}{\placeholder}   & \multirow{2}{*}{\thead{$\langle dla,2112,40 \rangle$ \\ $\langle shi, 1184, 24 \rangle$}} & CIFAR-10 &  92.62\% &  {\textbf{6.48e5}} & {\textbf{2.50e9}} & {\textbf{3.34e9}}\\
    & &  & STL-10   &  75.72\% & {$\checkmark$} & {$\checkmark$} & {$\checkmark$}  \\
    \hline 
    \multicolumn{4}{l}{$\times$: violate design specs;} & \multicolumn{4}{l}{{$\checkmark$} : meet design specs.}\\
    \hline
  \end{tabular}
\end{table}

Table \ref{tab:resMulti} reports the comparison results on multi-dataset workloads.
We implement two additional approaches.
First, ``NAS$\rightarrow$ASIC'' indicates successive NAS \cite{zoph2016neural} and brute-force hardware exploration.
Second, in ``ASIC$\rightarrow$HW-NAS'', a Monte Carlo search with 10,000 runs is first conducted to obtain the ASIC design closest 
to the design specs.
Then, for that specific ASIC design, we extend the hardware-aware NAS \cite{tan2018mnasnet} to identify the best neural architecture under the design specs.


Results in Table \ref{tab:resMulti} demonstrate that for the neural architectures identified by NAS, none of the accelerator designs explored by the brute-force approach can provide a legal solution that satisfies 
all design specs.
On the contrary, for both workloads, \placeholder~can guarantee the solutions to meet all specs with the average accuracy loss of 0.76\% and 1.17\%, respectively. 
For workload $W1$, \placeholder~achieves 17.77\%, 2.49$\times$, and 2.32$\times$ reductions on latency, energy, and area, respectively, against NAS$\rightarrow$ASIC.
For workload $W2$, the numbers 
are 30.39\%, 29.58\%, and 30.85\%.
When comparing \placeholder~with ASIC$\rightarrow$HW-NAS, even though the solution of the latter is closer to the design specs, for W1,  \placeholder~achieves 0.87\% higher accuracy for CIFAR-10 and similar accuracy for Nuclei; for W2, 3.65\% higher accuracy is achieved for STL-10 and similar accuracy for CIFAR-10. 



All the above results have revealed the necessity and underscored the importance of co-exploring neural architectures and ASIC designs.

    
    
    


\noindent\textit{D. From Single and Homogeneous to Heterogeneous ASIC Accelerator}

The benefits of heterogeneous accelerators under heterogeneous 
workloads are evident. 
Table \ref{tab:cifarExp} reports the comparison results of different accelerator configurations under the homogeneous workload CIFAR-10 (W3).
In these approaches, ``NAS'' explores neural architectures without hardware awareness and the corresponding ASIC applies the maximum hardware resource;
``Single Acc.'', ``Homo. Acc.'', ``Hetero. Acc'' are  \placeholder~with single accelerator design, two homogeneous sub-acclerators, and two heterogeneous sub-accelerators. 
Kindly note that, as discussed in Section \ref{s:Experiment}-A, \placeholder~can support the exploration of a single accelerator. We set hardware configurations as follows to guarantee single and homongeneous solutions to meet design specs.
For Single Acc., the network will be sequentially executed twice, which indicates that the constraints on latency and energy should be halved.
For Homo. Acc., two homogeneous sub-accelerators will run a same network simultaneously, which indicates that the energy and area for each accelerator should be halved.

From the results in Table \ref{tab:cifarExp}, we observe that although NAS can successfully identify the neural architectures with the highest accuracy (94.17\%), they cannot satisfy the specs even though all hardware resources are used.
In comparison, Single Acc. identifies a relatively smaller neural architecture with less hardware resource, but can meet the specs with the accuracy of 91.45\%.
Without exploring parallelism, Single Acc. cannot further improve accuracy since it is bounded by latency.
After boosting performance, Homo. Acc. identifies the neural architecture with 92.00\% accuracy.
Exploring the heterogeneous accelerators by \placeholder, two distinct networks are generated: one is with accuracy of 93.23\%, close to the best result identified by NAS; and the other one with slightly lower accuracy of 91.11\% is comparable with that of Single Acc..
This solution will be useful in Ensemble learning \cite{perrone1992networks}, and can provide more choices for designers.

\begin{table}[t]
  \scriptsize
  \centering
  \tabcolsep 1.7pt
  \renewcommand\arraystretch{1.3}
  \caption{On CIFAR-10 (W3), comparison results of architectures and accelerator designs obtained by different accelerator configurations.}
  \label{tab:cifarExp}
  \begin{tabular}{c|c|c|c|c}
    \hline
     Approach    & Hardware & Architecture  &  Accuracy & Sat. \\
    \hline 
     NAS & $\langle dla,4096,64 \rangle$ & $\langle 32,128,2,256,2,256,2 \rangle$  &  94.17\% & $\times$ \\
    \hline
     Single Acc. & $\langle dla,3104,24 \rangle$   & $\langle 8,32,2,128,1,256,1 \rangle$ &  91.45\% & \checkmark  \\
     
    \hline
     Homo. Acc. &  $2\times \langle dla,1408,32 \rangle$& $2\times \langle 32,32,1,128,1,256,1 \rangle$ &  92.00\% & \checkmark \\
    
    \hline
    Hetero. Acc.   & \multirow{2}{*}{\thead{$\langle dla,1760,56 \rangle$ \\ $\langle shi,1152,8 \rangle$}} & $\langle 8, 64,2,256,2,256,2 \rangle$ &  \textbf{93.23\%} & \multirow{2}{*}{\checkmark }\\
    (\placeholder)&   & $\langle 8, 32,2,128,2,128,1 \rangle$ &  91.11\% \\
    \hline 
    \multicolumn{5}{l}{$\langle FN_0,FN_1,Sk_1,FN_2,Sk_2,FN_3,SK_3 \rangle$: For the $i^{th}$ block, $FN_i$ is filter }\\
    \multicolumn{5}{l}{ numbers, $SK_i$ is skip layer numbers. Block 0 is a standard conv instead of residual.} \\
    \hline

  \end{tabular}
\end{table}

\section{Conclusion}
\label{s:Conclusion}

In this work, we have proposed a framework, namely \placeholder, to co-explore neural architectures and ASIC accelerator designs targeting multiple AI tasks on edges devices. \placeholder~has filled the missing link between NAS and ASIC by creating an accelerator template set in terms of the dataflow style. 
In addition, a novel multi-task oriented RNN controller has been developed to simultaneously determine multiple neural architectures under a unified design spec. 
The efficacy of \placeholder~is verified through a set of comprehensive experiments.



\bibliographystyle{unsrt}
{\small
\bibliography{Ref}
}
\end{document}